\documentclass[conference,a4paper]{IEEEtran}
\IEEEoverridecommandlockouts

\usepackage[hidelinks]{hyperref}
\usepackage[cmex10]{amsmath}
\usepackage{amssymb,amsfonts}
\usepackage{dblfloatfix}
\usepackage{multirow}
\usepackage{subfloat}
\usepackage[ruled,vlined]{algorithm2e}
\usepackage{graphicx}
\graphicspath{{Figures/PDF/}{Figures/PNG/}}
\usepackage{caption}
\usepackage{subfig}

\usepackage{booktabs}
\usepackage{siunitx}
\usepackage[numbers,compress]{natbib}
\usepackage{texnames}
\usepackage{bm,bbm}
\usepackage{orcidlink}
\usepackage{balance}

\makeatletter
\usepackage[a4paper, top=1.7cm, bottom=1.6cm, left=1.1cm, right=1.1cm]{geometry}
\makeatother

\begin{document}

\title{DDUNet: Dual Dynamic U-Net for Highly-Efficient Cloud Segmentation%
\thanks{This research was conducted with the financial support of Science Foundation Ireland under Grant Agreement No.\ 13/RC/2106\_P2 at the ADAPT SFI Research Centre at University College Dublin. The ADAPT Centre for Digital Content Technology is partially supported by the SFI Research Centres Programme (Grant 13/RC/2106\_P2) and is co-funded under the European Regional Development Fund.}
\thanks{Send correspondence to \mbox{S.\ Dev, E-mail: soumyabrata.dev@ucd.ie.}}}

\author{%
    \IEEEauthorblockN{Yijie~Li\orcidlink{0000-0003-2118-2280}$^{1}$, Hewei~Wang\orcidlink{0000-0002-6952-0886}$^{2}$, Jinfeng~Xu\orcidlink{0009-0001-7876-3740}$^{3}$, Puzhen~Wu\orcidlink{0000-0003-1510-215X}$^{4}$, Yunzhong~Xiao\orcidlink{0000-0001-6875-7649}$^{2}$, Shaofan~Wang\orcidlink{0000-0002-3045-624X}$^{5}$, Soumyabrata~Dev\orcidlink{0000-0002-0153-1095}$^{4,6}$}
    \IEEEauthorblockA{%
        $^{1}$Northwestern University
        $^{2}$Carnegie Mellon University
        $^{3}$The University of Hong Kong \\
        $^{4}$University College Dublin
        $^{5}$Beijing University of Technology
        $^{6}$The ADAPT SFI Research Centre}
}

\maketitle
\begin{abstract}
Cloud segmentation amounts to separating cloud pixels from non-cloud pixels in an image. Current deep learning methods for cloud segmentation suffer from three issues. (a) Constrain on their receptive field due to the fixed size of the convolution kernel. (b) Lack of robustness towards different scenarios. (c) Requirement of a large number of parameters and limitations for real-time implementation. To address these issues, we propose a Dual Dynamic U-Net (DDUNet) for supervised cloud segmentation. The DDUNet adheres to a U-Net architecture and integrates two crucial modules: the dynamic multi-scale convolution (DMSC), improving merging features under different reception fields, and the dynamic weights and bias generator (DWBG) in classification layers to enhance generalization ability. More importantly, owing to the use of depth-wise convolution, the DDUNet is a lightweight network that can achieve 95.3\% accuracy on the SWINySEG dataset with only 0.33M parameters, and achieve superior performance over three different configurations of the SWINySEg dataset in both accuracy and efficiency. Our code is publicly available at: \url{https://github.com/Att100/DDUNet}.
\end{abstract}


\section{Introduction}
Cloud information analysis is necessary and important for meteorology research. The distribution or form of the cloud can reflect specific information that can be used to learn the weather and generate advanced predictions. Generally, cloud images are taken by the meteorological satellite in the near-earth orbit, but in recent years, ground-based sky cameras~\cite{jain2021extremely,dev2015design} have been widely used because of their better temporal and spatial resolutions. Several datasets of optical RGB images captured by these sky cameras are released to the community, including SWIMSEG \cite{dev2016color}, SWINSEG \cite{dev2017nighttime}, and SWINySEG \cite{dev2019cloudsegnet}. With the development of deep neural networks, cloud segmentation for meteorology was further developed. A great number of full convolution network (FCN) \cite{long2015fully} and feature pyramid network (FPN) \cite{lin2017feature} based structures are used for cloud segmentation which consists of a backbone encoder and series of specially designed decoders. However, in recent years, with the development of mobile devices and embedded systems, there is a growing need for a lightweight and efficient model that can perform the segmentation in real-time on those devices. Many previous works have excellent performances, but they usually have large model sizes which makes it hard to perform real-time inference.

In this paper, we introduce Dual Dynamic U-Net (DDUNet) which uses U-Net as fundamental architecture and we proposed Dynamic Multi-scale Convolution (DMSC) in which multiple depth-wise convolutions with different dilation rates are adopted to increase the reception field and feature extraction ability without too many parameters. We also introduce Dynamic Wights and Bias Generator (DWBG) for our decoders to improve the generalization ability. We evaluate DDUNet on three different configurations, day-time, night-time, and day+night time of the SWINySEG dataset that confirm its effectiveness.

\section{Related Works}
Cloud image segmentation methods can be broadly categorized into traditional approaches~\cite{long2006retrieving, dev2014systematic} and deep learning methods~\cite{dev2019cloudsegnet, dev2019multi}. Traditional methods, such as those by Dev \textit{et al.}~\cite{dev2014systematic}, utilize color features, fixed convolution filters, and PCA with fuzzy clustering to highlight color differences between clouds and the sky. While effective in capturing overall distributions, these methods often lack detail, leading to lower segmentation accuracy. Deep learning approaches have significantly improved segmentation performance. Dev \textit{et al.} \cite{dev2019cloudsegnet} introduced CloudSegNet, an FCN-based method for binary cloud masks, and later expanded to multi-label segmentation\cite{dev2019multi}, classifying images into thin cloud, thick cloud, and sky. Shi \textit{et al.} \cite{shi2020cloudu} proposed CloudU-Net, combining U-Net with CRFs for refined segmentation, and enhanced it with dilated convolutions for a larger receptive field. CloudU-NetV2 \cite{shi2021cloudu} improved spatial and channel feature optimization using attention mechanisms and employed the RAdam optimizer for better convergence. Recently, Li \textit{et al.}~\cite{li2024ucloudnet} introduced UCloudNet, leveraging U-Net with residual connections and deep supervision for enhanced training. Recent research in remote sensing, such as superpixel-based methods for clustering hyperspectral images~\cite{cui2024superpixel} and real-time analysis of UAV imagery~\cite{cui2024real}, also underscores the importance of lightweight and scalable architectures in real-world applications. Recent research in remote sensing, such as superpixel-based methods for clustering hyperspectral images~\cite{cui2024superpixel} and real-time analysis of UAV imagery~\cite{cui2024real}, also underscores the importance of lightweight and scalable learning-based model architectures in real-world applications.

\begin{figure*}[!h]
 \raggedright
 \subfloat[The architecture of the DDUNet model. The structure of sub-modules has been omitted in this figure.]{
    \includegraphics[width=0.9\textwidth]{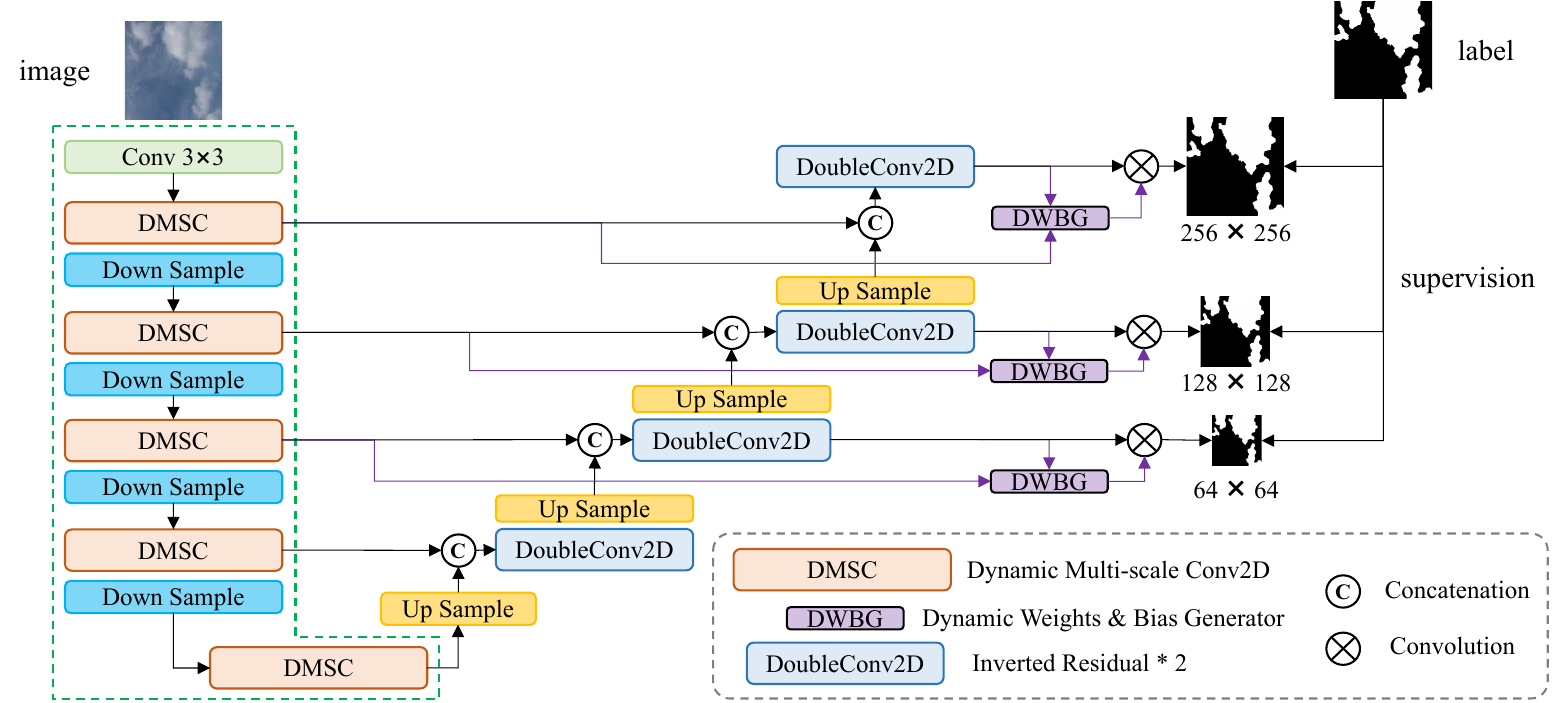}
    \label{fig:model-structure}
 }
 
 \subfloat[Dynamic Multi-scale Conv2D (DMSC), $d_1$, $d_2$, $d_3$, $d_4$ indicate the dilation rates of 1, 2, 3, 4. Linear-Layers is a group of layers constructed by (Linear, ReLU, and Linear).]{
    \includegraphics[width=0.55\textwidth]{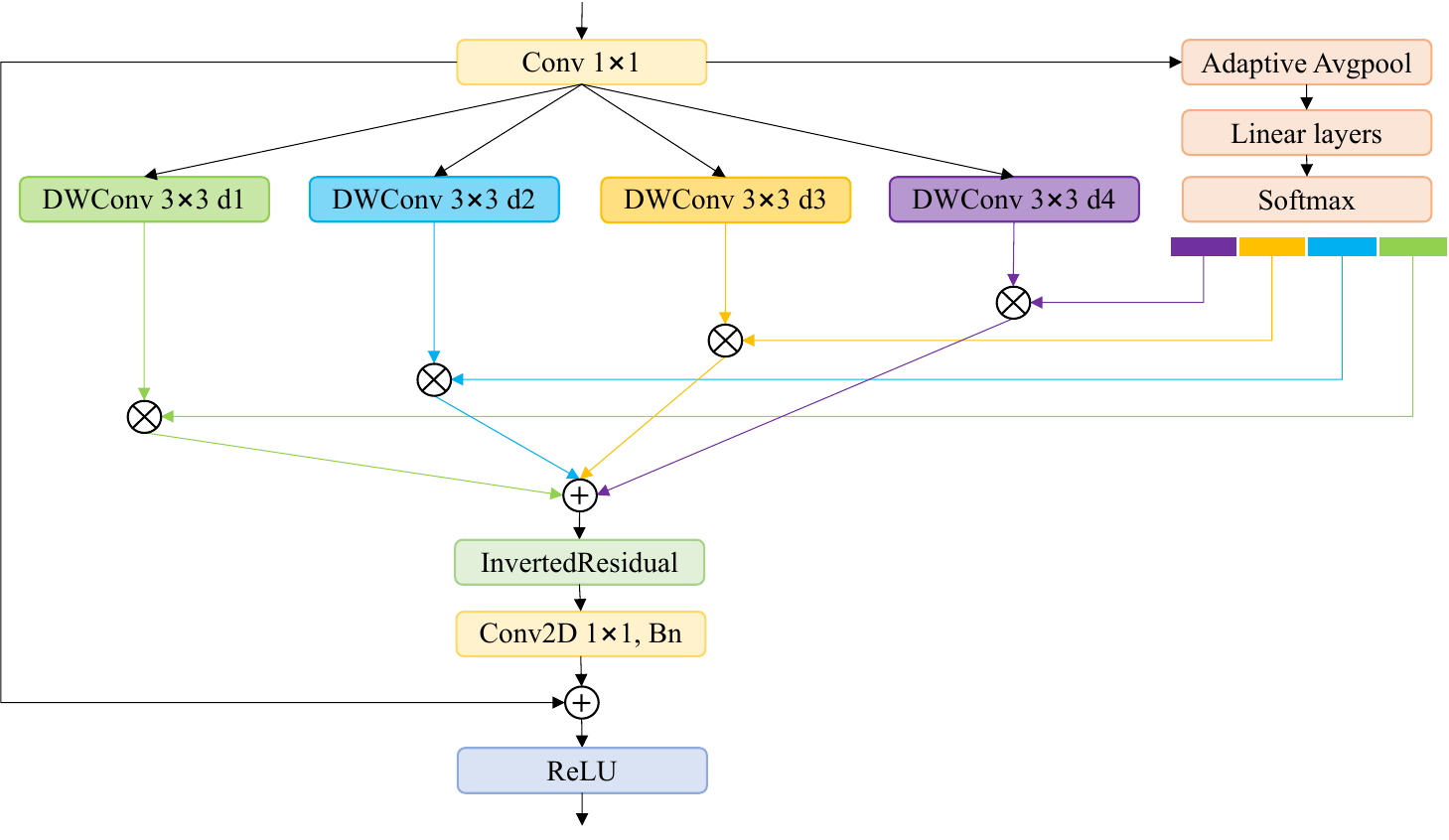}
    \label{fig:dmsc}
 }
 \hfill
 \subfloat[Dynamic Weights and Bias Generator (DWBG).]{
    \includegraphics[width=0.4\textwidth]{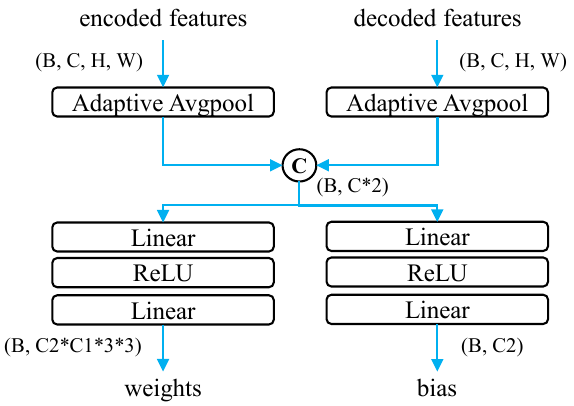}
    \label{fig-framework-c}
 }
 \caption{Overall Pipeline, DMSC, and DWBG.}
 \label{fig:dwbg}
 \vspace{-0.5cm}
\end{figure*}

\section{Architecture}
\label{sec:format}

DDUNet is built on the U-Net \cite{ronneberger2015u} architecture, consisting of a backbone encoder and four decoders with channel concatenation at each stage, as shown in Figure \ref{fig:model-structure}. Many computer vision works adopt encoder-decoder architectures integrated with CNNs to encode 2D image data~\cite{WANG2022102243, WANG2021, BATRA2022200039}. Encoder-decoder model architectures are widely applied in autonomous driving~\cite{WANGCAR2022}, medical imaging~\cite{tang2024optimized, pan2024accurate}, saliency object detection~\cite{li2023daanet, Li_2024_BMVC}, recommender system~\cite{xu2024aligngroup, xu2024mentor}, and robotics~\cite{zhenqi23, HuoAtten, zhu2023fanuc, wang2024airshot, wang2024onls}.  In our work, to enhance feature extraction efficiency, the encoder (green dashed area in Figure \ref{fig:model-structure}) includes four dynamic multi-scale convolution (DMSC) blocks and four stride-2 convolution layers, generating feature maps of sizes $32 \times 32$, $64 \times 64$, $128 \times 128$, and $256 \times 256$. In the decoder, four blocks progressively upsample feature maps from (H, W) to (2H, 2W) while reducing channels. Each block comprises two inverted residual \cite{sandler2018mobilenetv2} blocks and one upsample layer. Decoded feature maps pass through dynamic convolution layers with weights and biases generated by a dynamic weights and bias generator (DWBG). Finally, predictions from the last three stages are used for deep supervision to accelerate convergence.

\subsection{Basic Building Blocks}
The structures of these sub-modules are shown in 
Figure~\ref{fig:basic-building-blocks}. In Figure~\ref{fig:basic-building-blocks} (a), we use a depth-wise convolution layer, batch normalization layer, and a ReLU activation to construct a DWConv block for implementation of  Dynamic Multi-scale Conv2D (DMSC). The depth-convolution layer means the group of the convolution layer equals the number of input channels that take advantage of group convolution to reduce computation complexity. Figure~\ref{fig:basic-building-blocks} (b) shows the structure of the Conv block, we use both $1\times1$ filters and $3\times3$ filters in DDUNet. Figure~\ref{fig:basic-building-blocks} (c) shows the structure of Inverted Residual \cite{sandler2018mobilenetv2}. We use the Inverted Residual block in decoders to further reduce the inference time.

\subsection{Dynamic Multi-scale Conv2D (DMSC)}
We propose Dynamic Multi-Scale Conv2D (DMSC), shown in Figure~\ref{fig:dmsc}, to enhance multi-scale feature extraction by dynamically aggregating features of different scales. Traditional $1\times1$ or $3\times3$ convolutions have limited receptive fields, which hinder small object feature extraction. Dilated convolutions address this by introducing gaps (skip connections) to expand the receptive field. DMSC utilizes four dilation rates to extract features at varying scales. Similar to ASPP~\cite{chen2017deeplab}, which uses different dilation rates, and PSP~\cite{zhao2017pyramid}, which uses varied kernel sizes, DMSC improves context aggregation. In our design, a $1\times1$ Conv block maps input features into a new space, followed by five branches. The first branch applies adaptive average pooling and reshapes the feature map from (B, C, H, W) to (B, C, 1, 1) and then to (B, C),

\begin{equation}
    F_{\mathrm{c}} = \mathrm{Reshape}(\mathrm{AdptiveAvgPool}(F_{\mathrm{in}}))
\end{equation}
Then we use a series of linear layers to learn a weight vector that will be applied to the four multi-scale feature maps, which can be formulated as:

\begin{equation}
    F_{\mathrm{logits}} = \mathrm{Linear}(\mathrm{ReLU}(\mathrm{Linear}(F_{\mathrm{c}})))
\end{equation}
Apply softmax activation to retrieve the weight vector $W$,
\begin{equation}
    W = \mathrm{Softmax}(F_{\mathrm{logits}})
\end{equation}
After that, we apply the multi-scale dilated DWConvs to $F_{in}$, which can be formulated as:

\begin{equation}
    F_{\mathrm{d}}^{r} = \mathrm{DWConv_{3\times3}^{r}}(F_{\mathrm{in}}), r \in \{1, 2, 3, 4\}
\end{equation}
in which, $F_{\mathrm{d}}^{r}$ indicates the feature map after the convolution with a dilation rate of $r$. We then apply the weight vector $W$ to $\{F_{\mathrm{d}}^{1}, F_{\mathrm{d}}^{2}, F_{\mathrm{d}}^{3}, F_{\mathrm{d}}^{4}\}$, which can be given as,

\begin{equation}
    F_{\mathrm{a}} = \sum_{r=1}^{4} F_{\mathrm{d}}^{r} * W_{r}
\end{equation}
where $F_{\mathrm{a}}$ represents the aggregated feature map, $W_{r}$ indicates the weight element that will apply to the feature map $F_{\mathrm{d}}^{r}$. After that, we use a $3\times3$ Conv block to extract the feature, $F_{\mathrm{a}}^{'}$, which can be written as:

\begin{equation}
    F_{\mathrm{a}}^{'} = \mathrm{Conv_{3\times3}}(F_{\mathrm{a}})
\end{equation}
Finally, we apply a short-cut connection together with a $1\times1$ Conv, inspired by PSPNet \cite{zhao2017pyramid},

\begin{equation}
    F_{\mathrm{out}} = \mathrm{ReLU}(\mathrm{BN}(\mathrm{Conv2D_{1\times1}}(F_{\mathrm{a}})) + F_{\mathrm{in}})
\end{equation}

\textbf{\begin{figure}[htbp]
	\centering
	\includegraphics[height=1.7in]{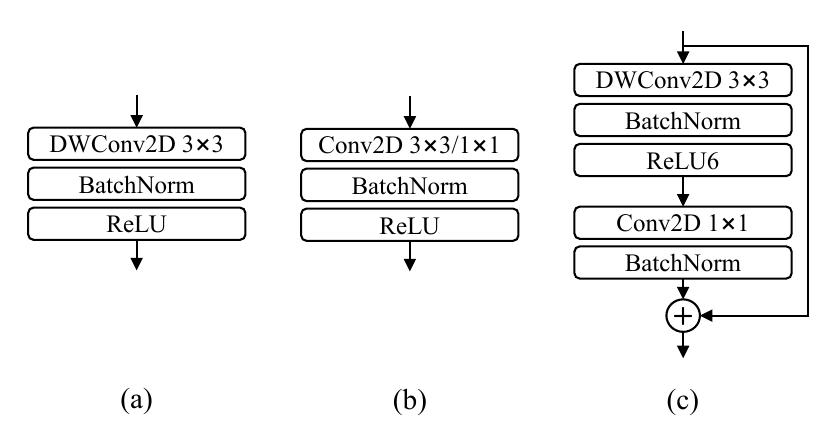}
	\caption{Basic building blocks used in DDUNet. (a) DWConv block with $3\times3$ filters; (b) Conv block with $1\times1$ or $3\times3$ filters; (c) Inverted Residual \cite{sandler2018mobilenetv2} without expand ratio.}\label{fig:basic-building-blocks}
\end{figure}}

\subsection{Dynamic Wights \& Bias Generator (DWBG)}
In most semantic segmentation models, a single Conv2D layer with fixed weights and biases is used for prediction, potentially limiting generalization. To address this, we implement a dynamic convolution layer with a dynamic weight \& bias generator (DWBG), as shown in Figure~\ref{fig:dwbg}. DWBG customizes weights for each input by processing encoder and decoder feature maps. These maps are squeezed using adaptive average pooling, concatenated along the channel axis, and reshaped to $(\mathrm{B}, \mathrm{C}*2)$. Two linear layers then generate weights and biases for the convolution operation. The generated weights have a shape of $(\mathrm{B}, \mathrm{C_{2}}, \mathrm{C_{1}}, 3, 3)$, where $\mathrm{C_{1}}$ and $\mathrm{C_{2}}$ are the input and output channels, respectively.

\subsection{Loss function}
We use binary cross entropy as the loss function and the total loss function can be represented as follows:

\vspace{-0.23in}
\begin{equation}
	\mathcal{L}_{\mathrm{bce}} = - \frac{1}{N} * \sum_{i=0}^{N} y_{i} * \log{p_{i}} + (1 - y_{i}) * \log{(1-p_{i})}
\end{equation} 
\vspace{-0.12in}
\begin{equation}
	\mathcal{L}_{\mathrm{total}} = \sum_{j=1}^{3} \alpha_{j} * \mathcal{L}_{\mathrm{bce}}^{j}
\end{equation}
in which, $p_{i}$ and $y_{i}$ indicate the $i$th pixel of the prediction and label. $N$ is the total number of pixels in the prediction map. $\alpha_{j}$ represent the weight of the loss value of $j$th decoder block. In our approach, we empirically set $\alpha_{1}=1$, $\alpha_{2}=0.5$ and $\alpha_{3}=0.2$.

\begin{figure*}[htbp]
	\centering
	\includegraphics[height=2.2in]{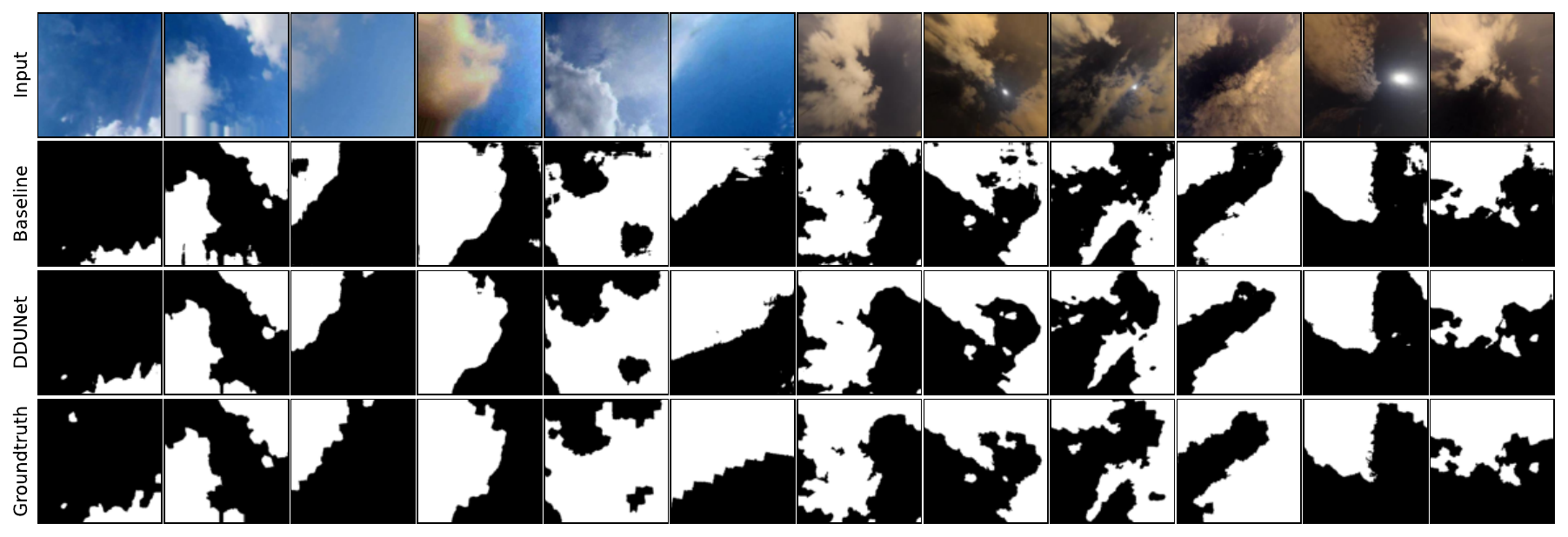}
	\caption{Results of cloud segmentation for day-time (1-6 columns) and night-time (7-12 columns).}\label{fig:Pred}
\end{figure*}

\tiny
\begin{table*}[htbp]
	\centering
	\caption{Comparison of DDUNet with other methods on SWINySEG dataset } \label{table-1}
	\scalebox{0.95}{
	\begin{tabular}{cccccccccccccc}
	    \cline{1-14}
		\multicolumn{1}{|c|}{\multirow{2}{*}{\textbf{Methods}}} & \multicolumn{1}{c|}{\textbf{\#Params}} & \multicolumn{4}{c|}{\textbf{Day-time}}  & \multicolumn{4}{c|}{\textbf{Night-time}} & \multicolumn{4}{c|}{\textbf{Day+Night time}}\\
		\multicolumn{1}{|c|}{} & \multicolumn{1}{c|}{(M)} & Acc. & Prec. & \scalebox{0.8}{$mF_{\beta}$} & \multicolumn{1}{c|}{MIoU} & Acc. & Prec. & \scalebox{0.8}{$mF_{\beta}$} & \multicolumn{1}{c|}{MIoU} & Acc. & Prec. & \scalebox{0.8}{$mF_{\beta}$} & \multicolumn{1}{c|}{MIoU}\\ 
		\cline{1-14}
		\multicolumn{14}{|c|}{General Semantic Segmentation Models} \\
		\cline{1-14}
		\multicolumn{1}{|c|}{U-Net \cite{ronneberger2015u}} & \multicolumn{1}{c|}{24.90} & .943 & .945 & .945 & \multicolumn{1}{c|}{.891} & .953 & .949 & .946 & \multicolumn{1}{c|}{.909} & .944 & .945 & .945 & \multicolumn{1}{c|}{.893}\\
		\multicolumn{1}{|c|}{PSPNet \cite{Zhao_2017_CVPR}} & \multicolumn{1}{c|}{6.92} & .945 & .953 & .948 & \multicolumn{1}{c|}{.896} & .938 & .927 & .929 & \multicolumn{1}{c|}{.882} & .945 & .951 & .946 & \multicolumn{1}{c|}{.895}\\
		\multicolumn{1}{|c|}{DeeplabV3+ \cite{chen2017rethinking}} & \multicolumn{1}{c|}{2.753} & .953 & .962 & .955 & \multicolumn{1}{c|}{.911} & .947 & .931 & .939 & \multicolumn{1}{c|}{.898} & .953 & .960 & .954 & \multicolumn{1}{c|}{.910}\\
		\cline{1-14}
		\multicolumn{14}{|c|}{Special Designed Cloud Segmentation Models} \\
		\cline{1-14}
		\multicolumn{1}{|c|}{CloudSegNet \cite{dev2019cloudsegnet}} & \multicolumn{1}{c|}{0.005} & .893 & .888& .898 & \multicolumn{1}{c|}{.806} & .880 & .870 & .895 & \multicolumn{1}{c|}{.813} & .896 & .899 & .899 & \multicolumn{1}{c|}{.811}\\
		\multicolumn{1}{|c|}{SegCloud \cite{xie2020segcloud}} & \multicolumn{1}{c|}{19.61} & .941 & .953 & .943 & \multicolumn{1}{c|}{.889} & .955 & .936 & .948 & \multicolumn{1}{c|}{.912} & .942 & .952 & .944 & \multicolumn{1}{c|}{.891}\\
		\multicolumn{1}{|c|}{CloudU-Net \cite{shi2020cloudu}} & \multicolumn{1}{c|}{35.49} & .954 & .956 & .956 & \multicolumn{1}{c|}{.912} & .954 & .925 & .949 & \multicolumn{1}{c|}{.912} & .954 & .954 & .956 & \multicolumn{1}{c|}{.913}\\
        \multicolumn{1}{|c|}{CloudU-Netv2 \cite{shi2021cloudu}} & \multicolumn{1}{c|}{14.55} & .940 & .967 & .941 & \multicolumn{1}{c|}{.887} & .954 & .931 & .948 & \multicolumn{1}{c|}{.911} & .941 & .964 & .941 & \multicolumn{1}{c|}{.889}\\
        \multicolumn{1}{|c|}{MA-SegCloud \cite{zhang2022novel}} & \multicolumn{1}{c|}{16.3} & .969 & .971 & .970 & \multicolumn{1}{c|}{.940} & .969 & .960 & .965 & \multicolumn{1}{c|}{.940} & .969 & .970 & .970 & \multicolumn{1}{c|}{.940}\\
		\cline{1-14}
		\multicolumn{14}{|c|}{Proposed Model} \\
		\cline{1-14}
		\multicolumn{1}{|c|}{DDUNet} & \multicolumn{1}{c|}{0.33} & .953 & .953 & .948 & \multicolumn{1}{c|}{.882} & .954 & .951 & .940 & \multicolumn{1}{c|}{.900} & .953 & .952 & .947 & \multicolumn{1}{c|}{.884}\\
		\cline{1-14}
	\end{tabular}
    }
    \vspace{-0.5cm}
\label{quant}
\end{table*}
\normalsize

\section{Experiments \& Results}

\subsection{Experiments Setting}
We follow \cite{zhang2022novel} to split the SWINySEG dataset (contains $6078$ day-time cloud images and $690$ night-time cloud images) with a ratio of 9:1 for training and testing. The batch size is 16 and the models are trained for 100 epochs. We use Adam with an initialized learning rate of 1e-3. Exponential learning-rate decay is applied with $\gamma$ equal to 0.95 after each training epoch. We evaluate our model with four widely used metrics: accuracy, precision, F-measure, and MIoU.

\subsection{Qualitative Analysis}
We compare the predictions of the baseline model (lightweight U-Net, 0.32M parameters), DDUNet, and ground truth, as shown in Figure \ref{fig:Pred}. The first six columns are daytime images, and the last six are nighttime. DDUNet produces more complete masks (e.g., columns 3, 6, and 11), correctly segmenting large patches missed by the baseline. It also performs better on small cloud patches (e.g., columns 1 and 3) and reduces false positives in large cloud-labeled areas, improving overall segmentation accuracy.

\subsection{Quantitative Analysis}
Quantitative evaluation results of our methods are shown in Table \ref{quant}, which shows the accuracy, precision, F-measure, and MIoU of DDUNet with other methods on day-time, night-time, and day+night time images. DDUNet only has 0.33M parameters which enables the ability to run in very small latency on computational resources-limited devices. By comparing with the current smallest model, the CloudSegNet \cite{dev2019cloudsegnet}, with only 0.005M parameters achieves 89.6\% accuracy on day+night time split and DDUNet has more parameters, but it can achieve 95.3\% accuracy on the same split. Because the parameter amount is under 0.5M, the difference in inference latency between the two models can be ignored and the advancement of DDUNet is highlighted. Besides, those methods that have similar performance with DDUNet, such as DeeplabV3+ \cite{chen2017rethinking} and CloudU-Net \cite{shi2020cloudu} can both achieve 95\% accuracy but the parameters number of DDUNet is only 1/9 and 1/100 of them.

\subsection{Ablation Study}
We perform an ablation study on model components and parameter sizes, as summarized in Table \ref{ablation_result}. A lightweight U-Net baseline with 0.32M parameters achieves 93.0\% accuracy and an MIoU of 0.839. Replacing the baseline encoder with DMSCs reduces parameters to 0.28M while improving accuracy to 94.8\% and MIoU to 0.873. Adding DWBGs further boosts DDUNet’s performance to 95.3\% accuracy and 0.884 MIoU. We also vary model size using the hyperparameter $base\_channels$ (set to 8 in prior experiments). Reducing it to 4 results in a smaller model, while increasing it to 16 yields limited performance gains, confirming that $base\_channels = 8$ achieves an optimal balance between performance and size.

\begin{table}[htbp]
	\centering
    \scalebox{0.85}{
	\begin{tabular}{c|c|c|cccc}
		\toprule
		\multirow{2}{*}{\textbf{No.}} & \multirow{2}{*}{\textbf{Methods}} & \multirow{2}{*}{\textbf{\#Params}} & \multicolumn{4}{c}{\textbf{SWINySEG}} \\ \cmidrule(lr){4-7}
		& & & Acc.  & Prec. & $mF_{\beta}$ & MIoU \\ 
		\midrule
		1 & Baseline & 0.32M & .930  & .937  & .925  & .839 \\
		2 & No. 1+DMSC & 0.28M & .948  & .943  & .942  & .873 \\
		3 & No. 2+DWBG & 0.33M & .953  & .952  & .947  & .884 \\
		5 & No. 3 (0.27x) & \textbf{0.09M} & .936 & .933 & .932  & .849 \\
		6 & No. 3 (3.79x) & 1.25M & \textbf{.954} & \textbf{.955}  & \textbf{.947} & \textbf{.886} \\ 
		\bottomrule
	\end{tabular}
    }
	\caption{Ablation study on different module compositions and different scaling on parameters amount}
	\label{ablation_result}
    \vspace{-0.5cm}
\end{table}

\section{Conclusion and Discussion}
In this paper, we introduce the dual dynamic U-Net (DDUNet) for cloud segmentation aiming to achieve a balance between accuracy and efficiency. By introducing dynamic multi-scale convolution (DMSC), DDUNet can extract the features of cloud patches with different scales and adaptive merge the feature maps. The dynamic weights and bias generator (DWBG) adaptive generates weights and bias for the final classification layer which enhances generalization ability across various scenarios. Significantly, its use of depth-wise convolution renders the DDUNet lightweight, achieving high accuracy (95.3\%) on the SWINySEG dataset with minimal parameters (0.33M). 
\balance 

\bibliographystyle{IEEEtran.bst}
\bibliography{strings,refs}

\end{document}